\documentclass{article}
\usepackage{ijcai18}

\usepackage{times}
\usepackage{xcolor}
\usepackage{graphicx}
\usepackage{amsmath}
\usepackage{amssymb}
\DeclareMathOperator*{\argmin}{argmin} 
\usepackage{bbding}
\usepackage{soul}
\usepackage[utf8]{inputenc}
\usepackage[small]{caption}
\usepackage{overpic}
\usepackage{booktabs}
\usepackage{diagbox}
\usepackage{etoolbox}
\apptocmd{\sloppy}{\hbadness 10000\relax}{}{}
\usepackage[multiple]{footmisc}
\definecolor{hyperref-green}{RGB}{0,150,0}
\definecolor{hyperref-red}{RGB}{200,0,0}
\usepackage[hidelinks]{hyperref}

\usepackage{fancyhdr}
\newcommand{\figref}[1]{Fig.~\ref{#1}}%
\newcommand{\tabref}[1]{Tab.~\ref{#1}}%

\renewcommand{\eqref}[1]{Eq.~(\ref{#1})}
\newcommand{\etal}[0]{{\emph{et. al.}}}%

\title{Hi-Fi: Hierarchical Feature Integration for Skeleton Detection}
\author{
Kai Zhao$^1$,
Wei Shen$^2$,
Shanghua Gao$^1$,
Dandan Li$^2$,
Ming-Ming Cheng$^1$\thanks{M.M. Cheng is the corresponding author.}\\
\\
$^1$ College of Computer and Control Engineering, Nankai University\quad \quad \\
$^2$ Key Laboratory of Specialty Fiber Optics and Optical Access Networks, Shanghai University\\
{\tt\small \url{http://mmcheng.net/hifi/}}
}

\begin{document}

\maketitle

\begin{abstract}
In natural images, the scales (thickness) of object skeletons may
dramatically vary among objects and object parts,
making object skeleton detection a challenging problem.
We present a new convolutional neural network (CNN) architecture
by introducing a novel hierarchical feature integration mechanism,
named Hi-Fi, to address the skeleton detection problem.
The proposed CNN-based approach has a powerful multi-scale 
feature integration ability that intrinsically captures 
high-level semantics from deeper layers
as well as low-level details from shallower layers.
By hierarchically integrating different CNN feature levels 
with bidirectional guidance,
our approach 
(1) enables mutual refinement across features of different levels, and 
(2) possesses the strong ability to capture both rich object context
and high-resolution details.
Experimental results show that our method
significantly outperforms the state-of-the-art methods
in terms of effectively fusing features from very different scales,
as evidenced by a considerable performance improvement on several benchmarks.
Code is available at \urlstyle{rm}\url{http://mmcheng.net/hifi}.
\end{abstract}

\section{Introduction}
Object skeletons are defined as the medial axis of foreground objects surrounded
by closed boundaries~\cite{blum1967shape}.
Complementary to object boundaries, skeletons are shape-based descriptors
which provide a compact representation of both object geometry and topology.
Due to its wide applications in other vision tasks
such as shape-based image retrieval \cite{demirci2006object,sebastian2004recognition},
and human pose estimation 
\cite{girshick2011efficient,shotton2013real,sun2012conditional}.
Skeleton detection is extensively studied very recently
~\cite{shen2017deepskeleton,ke2017side,tsogkas2017amat}.

\begin{figure}[!t]
  \begin{overpic}[width=1\linewidth]{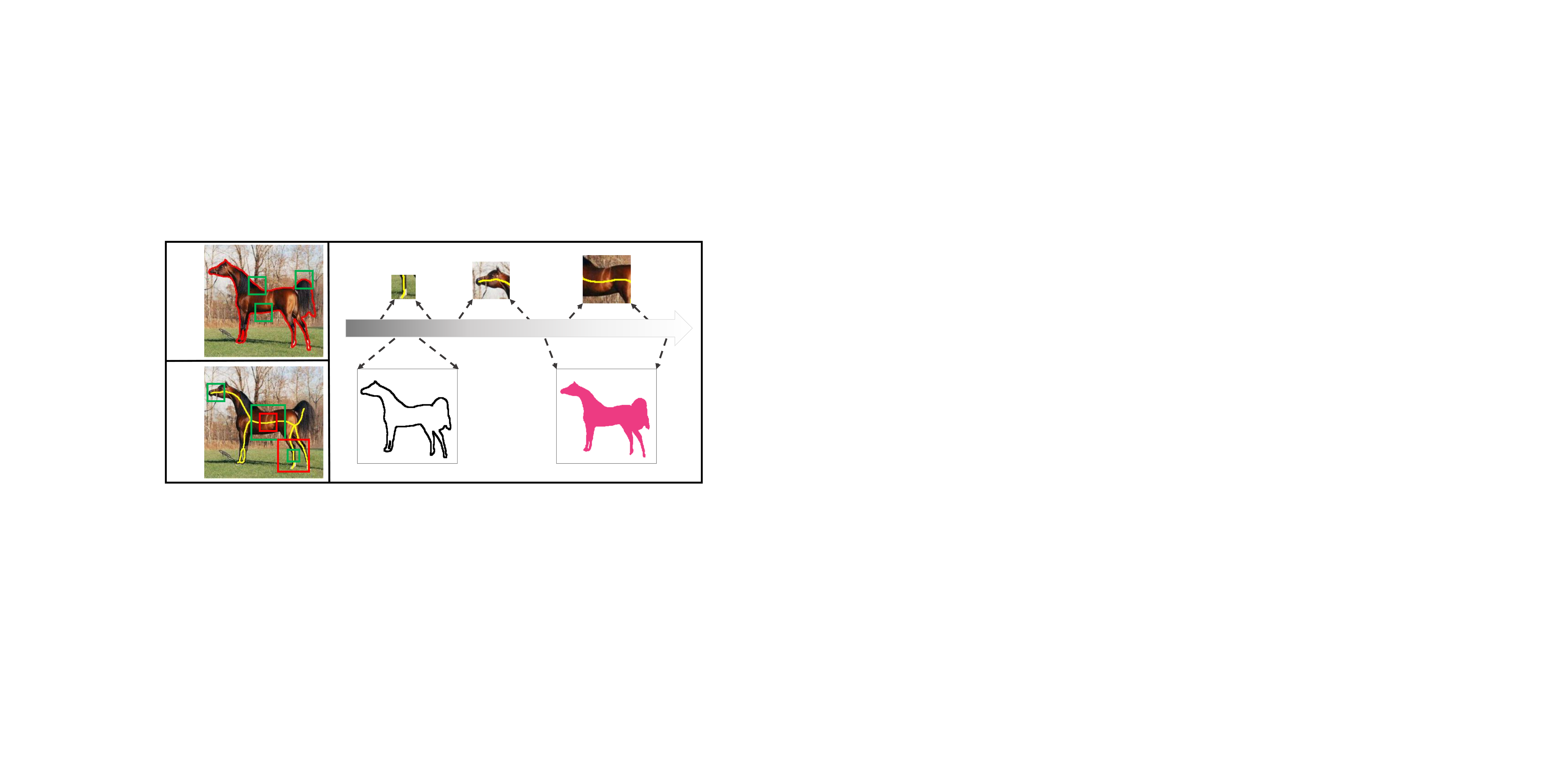}
  \put(2,10){(b)}
  \put(2,33){(a)}
  \put(62,10){(c)}
  \end{overpic}
  \vspace{-13pt}
  \caption{
  Skeleton detection is facing a more challenging scale space problem:
  (a) object boundaries can be detected with filters of
  similar size (green boxes);
  (b) only filters a bit larger than the skeleton scale (green boxes)
  can capture proper context for skeleton detection;
  both improper big or small (red boxes) cannot perceive skeletons well;
  (c) compared with boundary detection and semantic segmentation, 
  skeleton detection requires inhomogeneous feature levels.
  }\label{fig:boundary-sk}
  \vspace{-4pt}
\end{figure}

Because the skeleton scales (thickness) are unknown and 
may vary among objects and object parts,
skeleton detection has to deal with a more challenging 
scale space problem~\cite{shen2016object}
compared with boundary detection, as shown in \figref{fig:boundary-sk}.
Consequently, it requires the detector to capture broader context 
for detecting potential large-scale (thick) skeletons, 
and also possess the ability to focus on local details 
in case of small-scale (thin) skeletons.
\begin{figure*}[t]
\centering
  \begin{overpic}[width=0.9\linewidth]{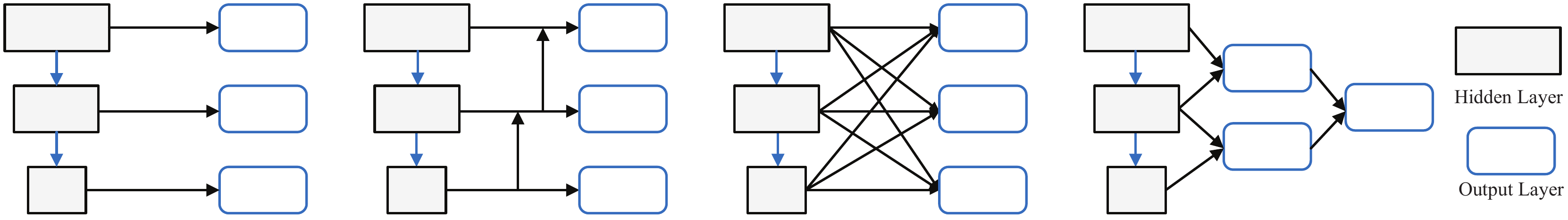}
  \put(9,-2){(a)}
  \put(32,-2){(b)}
  \put(56,-2){(c)}
  \put(77,-2){(d)}
  \end{overpic}
  \vspace{6pt}
  \caption{Different multi-scale CNN feature fusing methods:
    (a) side-outputs as independent detectors at different scales
    \protect \cite{xie2015holistically,shen2016object};
    (b) deep-to-shallow refinement \protect \cite{ke2017side} that brings
    high-level semantics to lower layers;
    (c) directly fuse all feature levels at once;
    (d) our hierarchical integration architecture,
    which enables bidirectional mutual refinement across
    low/high level features by recursive feature integration.}
  \label{fig:arch-feat-fuse}\vspace{-10pt}
\end{figure*}
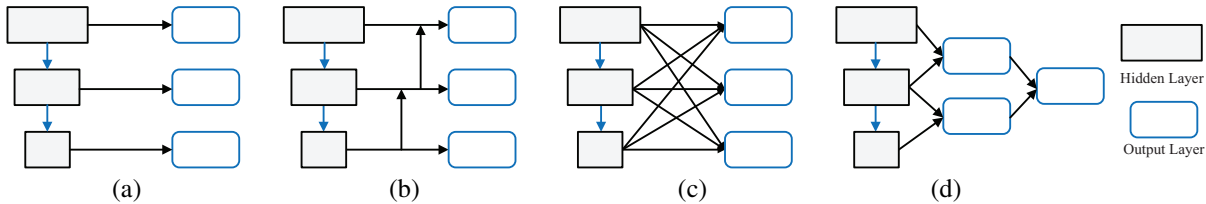

Performing multi-level feature fusion has been a primary trend in
pixel-wise dense prediction such as skeleton detection~\cite{ke2017side} 
and saliency detection~\cite{Zhang_2017_ICCV,hou2016deeply}.
These methods fuse CNN features of different levels 
in order to obtain more powerful representations.
The disadvantage of existing feature fusion methods is that 
they perform only deep-to-shallow refinement,
which provides shallow layers the ability of perceiving high-level concepts 
such as object and image background.
Deeper CNN features in these methods still suffer from low-resolution, 
which is a bottleneck to the final detection results.

In this paper we introduce hierarchical feature integration (Hi-Fi) 
mechanism with bidirectional guidance.
Different from existing feature fusing solutions, 
we explicitly enable both \emph{deep-to-shallow} and \emph{shallow-to-deep}
refinement to enhance shallower features with richer semantics,
and enrich deeper features with higher resolution information.
Our architecture has two major advantages compared with existing alternatives:

\paragraph{Bidirectional Mutual Refinement.}
Different from existing solutions illustrated in 
\figref{fig:arch-feat-fuse} (a) and \figref{fig:arch-feat-fuse} (b) 
where different feature levels work independently or allow only
\emph{deep-to-shallow} guidance, our method \figref{fig:arch-feat-fuse}
(d) enables not only \emph{deep-to-shallow} but also \emph{shallow-to-deep} refinement,
which allows the high-level semantics and low-level details
to mutually help each other in a bidirectional fashion.

\paragraph{Hierarchical Integration.}
There are two alternatives for mutual integration:
directly fusing all levels of features as shown in \figref{fig:arch-feat-fuse} (c),
or hierarchically integrating them as shown in \figref{fig:arch-feat-fuse} (d).
Due to the significant difference between faraway feature levels,
directly fusing all of them might be very difficult.
We take the second approach as inspired by the philosophy of ResNet~\cite{he2016deep},
which decomposes the difficult problem into much easier sub-problems:
identical mapping and residual learning.
We decompose the feature integration into easier sub-problems:
gradually combining nearby feature levels.
Because optimizing combinations of close features is more practical and easier to converge
due to their high similarity.
The advantage of hierarchical integration over directly fusing all feature levels
is verified in the experiments (\figref{fig:convergence}).

\section{Related Work}
\paragraph{Skeleton Detection.}
Numerous models have been proposed for skeleton detection in the past decades.
In general, they can be divided into three categories:
(a) early image processing based methods, these methods localize object skeleton
based on the geometric relationship between object skeletons and boundaries;
(b) learning based methods, by designing hand-crafted image features, these methods train a
machine learning model to distinguish skeleton pixels from non-skeleton pixels;
(c) recent CNN-based methods which design CNN architectures for skeleton detection.

Most of the early image processing based methods
~\cite{morse1993multiscale,jang2001pseudo}
rely on the hypothesis that skeletons lie in
the middle of two parallel boundaries.
A boundary response map is first calculated (mostly based on image gradient),
then skeleton pixels can be localized with the geometric relationship 
between skeletons and boundaries.
Some researchers then investigate learning-based models for skeleton detection.
They train a classifier \cite{tsogkas2012learning} or 
regressor \cite{sironi2014multiscale} with
hand-crafted features to determine whether a pixel touches the skeleton.
Boundary response is very sensitive to texture and illumination changes, 
therefore image processing based methods
can only deal with images with simple backgrounds.
Limited by the ability of traditional learning models and 
representation capacity of hand-crafted features,
they cannot handle objects with complex shapes and various skeleton scales.

More recently many researchers have been exploiting the powerful
convolutional neural networks (CNNs) for skeleton detection
and significant improvements have been achieved on several benchmarks.
HED~\cite{xie2015holistically} introduces \emph{side-output} 
that is branched from intermediate CNN
layers for multi-scale edge detection.
FSDS~\cite{shen2016object} then extends side-output to be \emph{scale-associated side-output},
in order to tackle the scale-unknown problem in skeleton detection.
The side-output residual network (SRN) \cite{ke2017side} exploits 
\emph{deep-to-shallow} residual connections to bring high-level, 
rich semantic features to shallower side-outputs with 
the purpose of making the shallower side-outputs more powerful 
to distinguish real object skeletons from local reflective structures.

\vspace{-10pt}\paragraph{Multi-Scale Feature Fusing in CNNs.}
CNNs naturally learn low/mid/high level features in a shallow to
deep layer fashion.
Low-level features focus more on local detailed structures,
while high-level features are rich in conceptual
semantics~\cite{zeiler2014visualizing}.
Pixel-wise dense prediction tasks such as skeleton detection,
boundary detection and saliency detection require
not only high-level semantics but also high-resolution predictions.
As pooling layers with strides down-sample the feature maps,
deeper CNN features with richer semantics are always with lower resolution.

Many researchers \cite{ke2017side,Zhang_2017_ICCV,hou2016deeply}
try to fuse deeper rich semantic CNN features with shallower high-resolution features
to overcome this \emph{semantic} vs \emph{resolution} conflict.
In SRN, Ke \etal~\shortcite{ke2017side} connected shallower side-outputs
with deeper ones to refine the shallower side-outputs.
As a result, the shallower side-outputs become much cleaner because they are 
capable of suppressing non-object textures and disturbances.
Shallower side-outputs of methods without deep-to-shallow refinement such as 
HED~\cite{xie2015holistically} and FSDS~\cite{shen2016object} are filled with noises.

A similar strategy has been exploited in DSS~\cite{hou2016deeply} and 
Amulet~\cite{Zhang_2017_ICCV} for saliency detection,
and a schema of these methods with \emph{deep-to-shallow} refinement 
can be summarized as \figref{fig:arch-feat-fuse} (b).
The problem of these feature fusion methods is that 
they lack the \emph{shallow-to-deep} refinement,
the deeper side-outputs still suffer from low-resolution.
For example, DSS has to \emph{empirically} drop the last side-output
for its low-resolution.

\begin{figure}[t]
  \centering
  \begin{overpic}[width=0.9\linewidth]{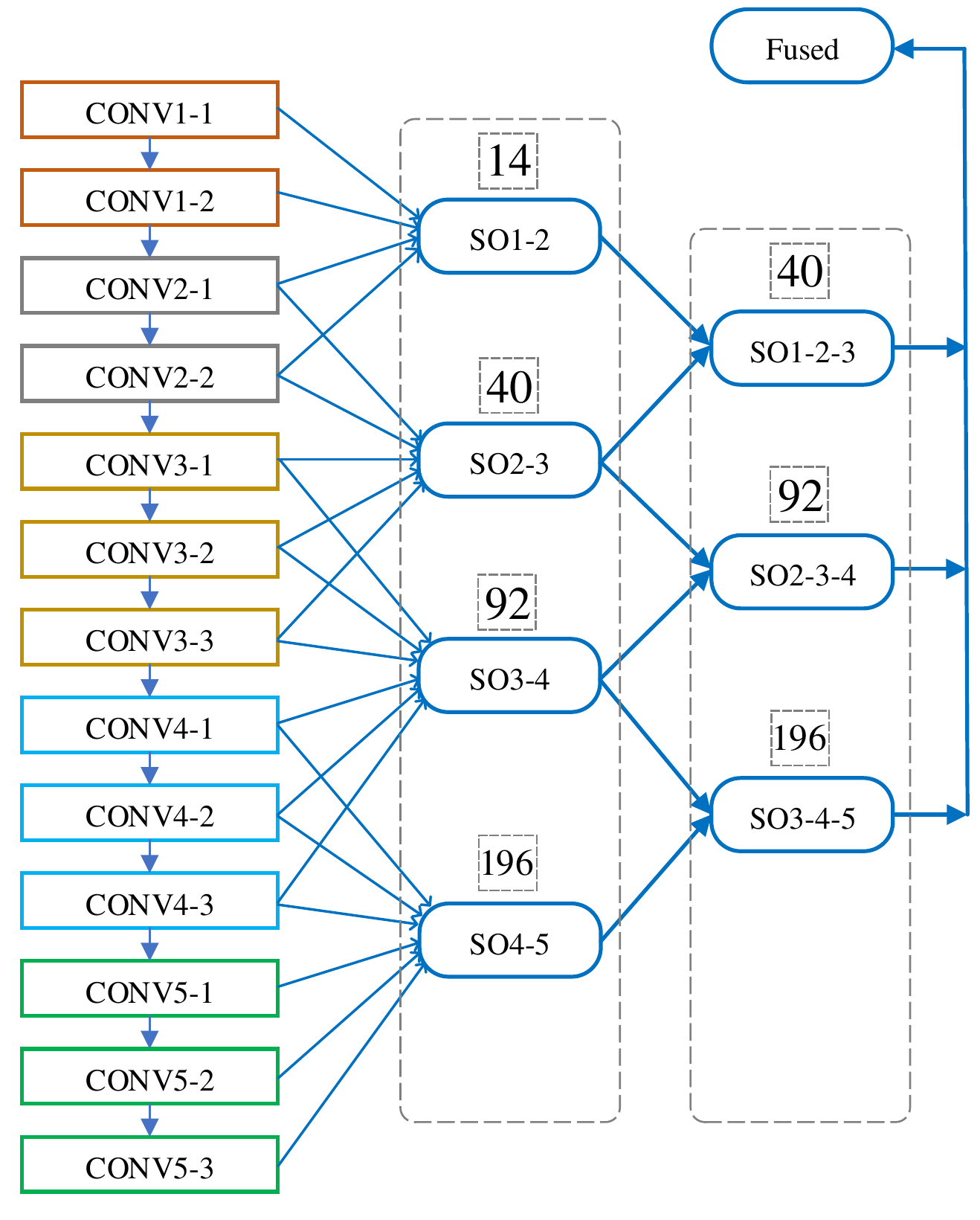}
   \put(37.0,10){Hi-Fi-1}
   \put(60.5,10){Hi-Fi-2}
  \end{overpic}
  \vspace{-8pt}
  \caption{Architecture of our proposed Hi-Fi network.
  	All side-outputs (SOs, marked with rounded square box) are supervised by skeletons
  	within their receptive fields (numbers on top of SOs indicate their receptive fields).
  	Features of neighbouring feature levels are integrated to enable mutual refinement,
  	and a lateral feature hierarchy is obtained with recursively 
  	integrating neighbouring features.
  }\label{fig:arch-fsds-srn-hifi2-h}
\end{figure}

\section{Hi-Fi: Hierarchical Feature Integration}
\subsection{Overall Architecture}
We implement the proposed Hi-Fi architecture based on the 
VGG16~\cite{simonyan2014very} network,
which has 13 convolutional layers and 2 fully connected layers.
The conv-layers in VGG network are divided into 5 groups: conv1-x, ..., conv5-x,
and there are 2$\sim$3 conv-layers in a group.
There are pooling layers with $stride=2$ between neighbouring convolution groups.

In HED, the side-outputs connect only with the last conv-layer of each group.
RCF (Richer Convolution Features)~\cite{liu2017richer} connects a side-output 
to all layers of a convolutional group.
We follow this idea to get more powerful convolutional features.
The overall architecture of Hi-Fi is illustrated in \figref{fig:arch-fsds-srn-hifi2-h}, 
convolutional groups are distinguished by colors, and pooling layers are omitted.

\subsection{Hierarchical Feature Integration} 
\begin{figure}[!htbp]
  \centering
  \includegraphics[width=\linewidth]{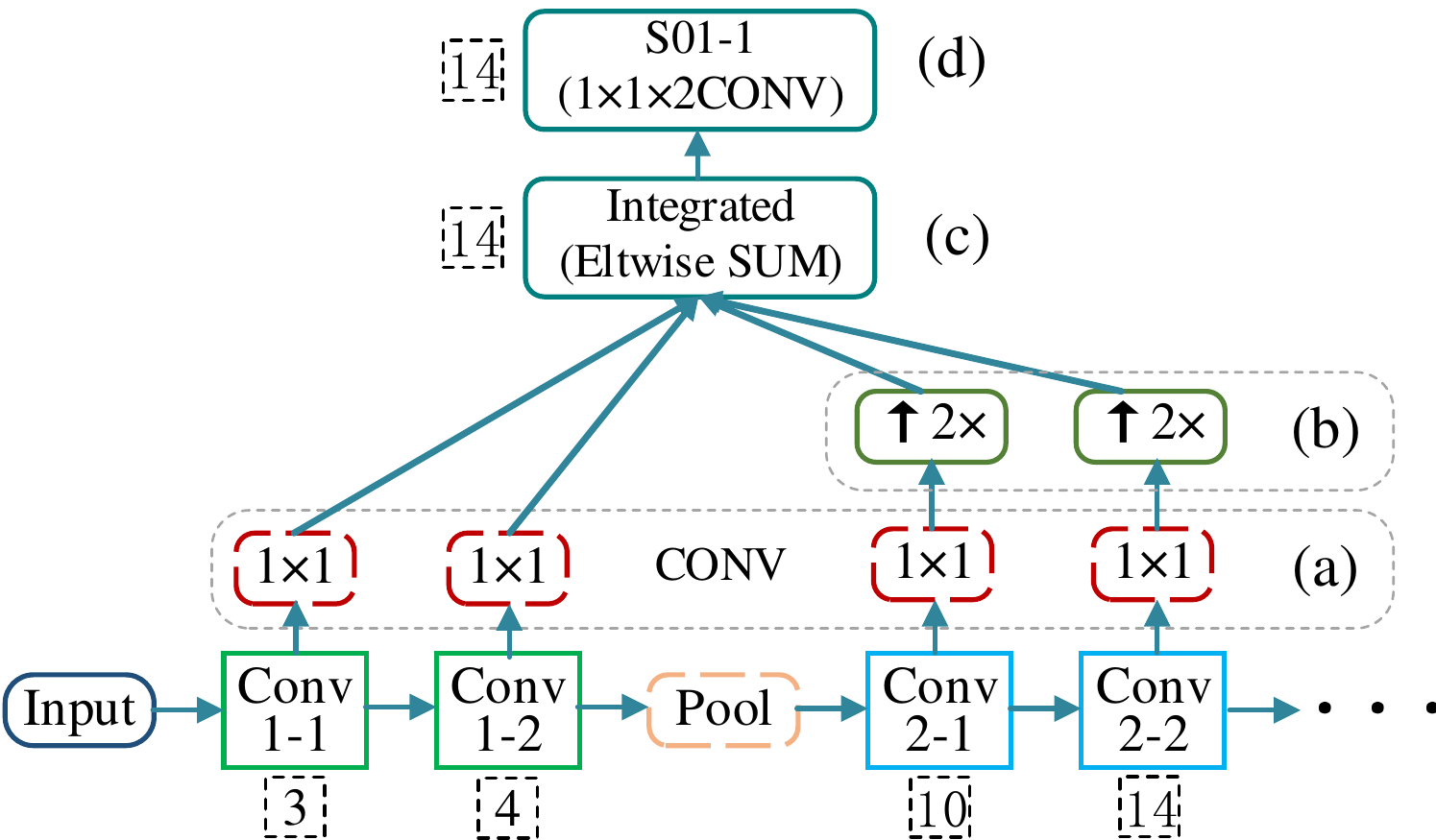}
  \vspace{-12pt}
  \caption{Illustration of feature integration:
  (a) feature maps to be integrated are firstly produced by a (1$\times$1) convolution;
  (b) deeper features are upsampled before integration;
  (c) the integration is implemented by an element-wise sum;
  (d) side-outputs are built on top of the integrated features 
  	with a (1$\times$1) convolution.
  }\label{fig:arch_integration}\vspace{-8pt}
\end{figure}

A detailed illustration of the proposed feature integration procedure is shown in
\figref{fig:arch_integration}.
Feature maps to be integrated are branched from
the primary network stream through a ($1 \times 1$) convolutional layer 
(dotted boxes marked with (a)) on top of an interior convolutional layer.
These feature maps are further integrated with element-wise sum (box marked with (c)).
The final scale-associated side-output (box marked with
(d)) is produced by a ($1 \times 1$) convolution.
Note that due to the existence of pooling layers,
deeper convolutional feature maps are spatially smaller than shallower ones.
Upsampling (box marked with (b)) is
required to guarantee all feature maps to be integrated are of the same size.

Ideally, the feature integration can be recursively performed until the last
integrated feature map contains information from all convolution
layers (conv1-1 $\sim$ conv5-3).
However, limited by the memory of our GPUs and the training time,
we end up with two level integration (\figref{fig:arch-fsds-srn-hifi2-h} `Hi-Fi-2').

\begin{figure}[t]
  \includegraphics[width=\linewidth]{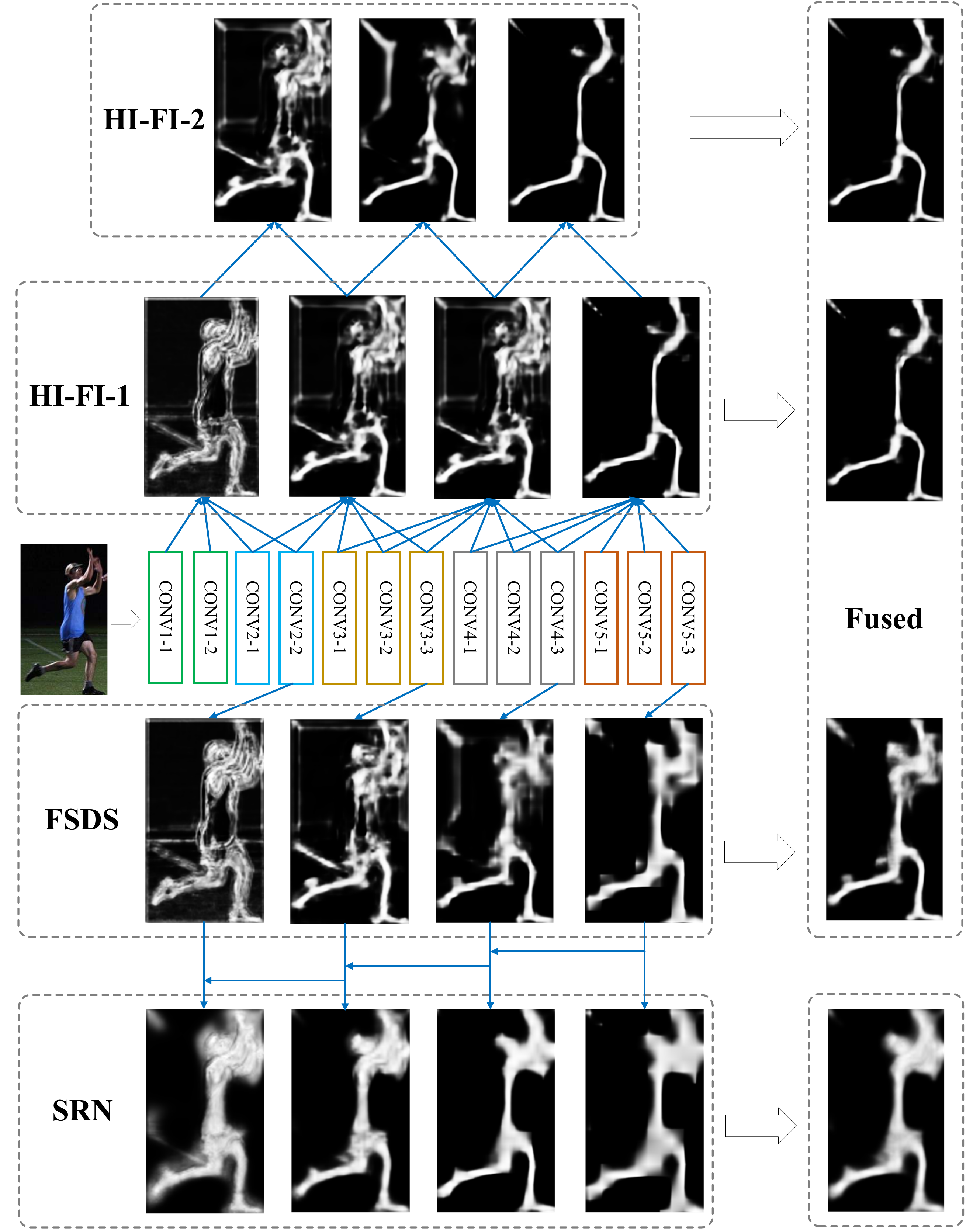}\\
  \vspace{-12pt}
  \caption{Illustration of Hi-Fi, FSDS \protect\cite{shen2016object} 
    and SRN \protect\cite{ke2017side}.
  }\vspace{-9pt}
  \label{fig:internal_results}
\end{figure}

\vspace{-10pt}\paragraph{Bidirectional Refinement.}
We explain the proposed \emph{bidirectional mutual refinement} by comparing it with existing
architectures: FSDS~\cite{shen2016object} and SRN~\cite{ke2017side}.
As shown in \figref{fig:internal_results}, 
side-outputs (SOs) of FSDS are working independently,
there is no cross talk between features of different levels.
As a result, FSDS has noisy shallow SOs and low-resolution deeper SOs.
SRN then introduces \emph{deep-to-shallow} refinement by 
bringing deep features to shallow SOs.
As shown in \figref{fig:internal_results}, 
shallower SOs of SRN are much cleaner than that of FSDS.
Despite the improvement, deeper SOs in SRN are still suffering from low-resolution, 
which limits the quality of the final fused result.

In our architecture SOs are  built on top of an integration of nearby feature levels,
and the ``nearby feature integration" is recursively performed.
In testing phase, SOs will receive information from both deeper and shallower sides;
and in training phase, gradient from SOs will back-prop to both as well.
In other words, our approach explicitly enables not only \emph{deep-to-shallow} but
also \emph{shallow-to-deep} refinement.
It is obviously shown in \figref{fig:internal_results} that Hi-Fi
obtains cleaner shallower SOs than FSDS, and at the same time
has much more high-resolution deeper SOs than SRN.
Consequently, we gain a strong quality improvement in the final fused result.

\subsection{Formulation}
Here we formulate our approach for skeleton detection.
Skeleton detection can be formulated as a pixel-wise binary classification problem.
Given an input image $X = \{x_j, j=1,...,|X|\}$, the goal of skeleton detection is
to predict the corresponding skeleton map $\hat{Y} = \{\hat{y_j}, j=1,...,|X|\}$,
where $\hat{y_j} \in \{0, 1\}$ is the predicted label of pixel $x_j$.
$\hat{y_j} = 1$ means pixel $x_j$ is predicted as a skeleton point, 
otherwise, pixel $x_j$ is the background.

\vspace{-12pt}\paragraph{Ground-truth Quantization.} 
Following FSDS~\cite{shen2016object}, we supervise the side-outputs
with `scale-associated' skeleton ground-truths.
To differentiate and supervise skeletons of different scales,
skeletons are quantized  into several classes according to their scales.
The skeleton scale is defined as the distance between a skeleton point 
and its nearest boundary point.
Assume $S=\{s_j, j=1,...,|X|\}(s_j \in R)$ is the skeleton scale map, where $s_j$
represents the skeleton scale of pixel $x_j$.  
When $x_j$ is the background, $s_j = 0$.
Let $Q=\{q_j, j=1,...,|X|\}$ be the quantized scale map, 
where $q_j$ is the quantized scale of pixel $x_j$.
The quantized scale $q_j$ can be obtained by:
\begin{equation}
  q_j = 
  \begin{cases}
  	m & \text{if $ r_{m-1} < s_j \leqslant r_{m}$}\\
	0 & \text{if $s_j = 0$ or $s_j > r_M$},
\end{cases}
\end{equation}
where  $r_m$ ($m=1,...,M$) is the receptive field of the $m$-th side-output (SO-$m$), 
with $r_0 = 0$, and $M$ is the number of side-outputs.
%
%
For instance, pixel $x_j$ with scale $s_j=39$ is quantized as $q_j = 3$, 
because $14 = r_2 < s_j \leqslant r_3=40$
(receptive fields of side-outputs are shown in 
\figref{fig:arch-fsds-srn-hifi2-h} with numbers-in-boxes).
All background pixels ($s_j=0$) and skeleton points out of scope of the network 
($s_j > r_M$) are quantized as 0.

\vspace{-10pt}\paragraph{Supervise the Side-outputs.}
Scale-associated side-output is only supervised by the skeleton 
with scale smaller than its receptive field.
We denote ground-truth as $G^m=\{g^m_j, j=1,...,|X|\}$, which is used to supervise SO-$m$.
$G^m$ is modified from $Q$ with all quantized values larger than $m$ set to zero.
$g^m_j$ can be obtained from $Q$ by:
\begin{equation}
  g^m_j = 
  \begin{cases}
	q_j &\text{if $q_j < m$}\\
	0 &\text{otherwise}.
  \end{cases}
  \label{eq:quant}
\end{equation}
Supervising side-outputs with the quantized skeleton map turns the original
binary classification problem into a multi-class problem.

\vspace{-10pt}\paragraph{Fuse the Side-outputs.}
Suppose $P^m = \{p^m_{j,k}, j=1,...,|X|, k=0,...,K^m \}$ 
is the predicted probability map of SO-$m$,
in which $K^m$ is the number of quantized classes SO-$m$ can recognise.
$p^m_{j,k}$ means the probability of pixel $x_j$ belonging to a quantized class \#$k$,
and index $j$ is over the spatial dimensions of the input image $X$.
Obviously we have $\sum_{k=0}^{K^m} p^m_{j, k} = 1$.

We fuse the probability maps from different side-outputs $P^m(m=1, ..., M)$
with a \textit{weighted summation} to obtain the final fused prediction $P=\{p_{j,k}\}$:
\begin{equation}\label{eq:fuse}
  p_{j,k} = \sum_{m} w_{mk} \cdot p^m_{j,k},
\end{equation}
where $p_{j,k}$ is the fused probability that pixel $x_j$ belongs to quantized class \#$k$,
$w_{mk}$ is the credibility of side-output $m$ on quantized class \#$k$.
\eqref{eq:fuse} can be implemented with a simple ($1 \times 1$) convolution.
%

\vspace{-10pt}\paragraph{Loss Function and Optimization.} 
We simultaneously optimize all the side-outputs $P^m(m=1, ..., M)$ 
and the fused prediction $P$ in an end-to-end way.
In HED~\shortcite{xie2015holistically}, 
Xie and Tu introduce a class-balancing weight to address 
the problem of positive/negative unbalancing in boundary detection.
This problem still exists in skeleton detection because most of the pixels are background.
We use the class-balancing weight to offset the imbalance between skeletons and the background.
Specifically, we define the balanced softmax loss $l(P, G)$ as:
\begin{equation}
\begin{split}
l(P, G) = \sum_j \Big[ -\beta^m \sum_{k \neq 0} \log(p^m_{j,k})1(g^m_j == k) - \\
  (1-\beta^m) \log(p^m_{j,0})1(g^m_j==0) \Big],
\end{split}
\label{eq:loss}
\end{equation}
where $P = h(X|\Theta)$ is the prediction from CNN,
$G$ is the ground-truth, and $m$ is the index of side-outputs.
$h(X|\Theta)$ is the model hypothesis taking image $X$ as input, parameterized by $\Theta$.
$\beta^m =\left[ \sum_{j=1}^{|X|} 1(g^m_j\neq0)\right]/|X|$ is a balancing factor, where $1(\cdot)$ is an indicator. The overall loss function can be expressed as follow:
\begin{equation}
\begin{split}
\mathcal{L} \big( h(X|\Theta), G \big) &= \mathcal{L}_{\text{side}} + \mathcal{L}_{\text{fuse}} \\
            &=\sum_{m=1}^{M} l(P^m, G^m) + l(P, Q),
\end{split}
\end{equation}
where $P^m$ and $G^m$ are prediction/ground-truth of SO-$m$ respectively, 
$G$ is the ground-truth of final fused output $P$.

All the parameters including the fusing weight $w_{mk}$ in \eqref{eq:fuse} 
are part of $\Theta$.
We can obtain the optimal parameters by a standard stochastic gradient descent (SGD):
\begin{equation}
(\Theta)^* = \argmin \mathcal{L}.
\end{equation}

\vspace{-10pt}\paragraph{Detect Skeleton with Pretrained Model.} 
Given trained parameters $\Theta$,
the skeleton response map $\hat{Y} = \{\hat{y}_j, j=1,...,|X|\}$ is obtained via:
\begin{equation}
\hat{y}_j = 1 - p_{j,0},
\label{eq:sk-resp}
\end{equation}
where $\hat{y}_j \in [0, 1]$ indicates the probability that pixel $x_j$ belongs to the skeleton.

\section{Experiments and Analysis}
In this section, we discuss the implementation details and report the
performance of the proposed method on several open benchmarks.
\subsection{Datasets}
The experiments are conducted on four popular skeleton datasets:
WH-SYMMAX~\cite{shen2016multiple}, SK-SMALL~\cite{shen2016object},
SK-LARGE\footnote{\urlstyle{rm}\url{http://kaiz.xyz/sk-large}}~\cite{shen2017deepskeleton} and SYM-PASCAL~\cite{ke2017side}.
Images of these datasets are selected from other semantic segmentation datasets
with human annotated object segmentation masks,
and the skeleton ground-truths are extracted
from segmentations.
Objects of SK-SMALL and SK-LARGE are cropped from MSCOCO dataset  with `well defined'
skeletons , and there is only one object in each image.
SYM-PSCAL selects images from the PASCAL-VOC2012 dataset without cropping 
and here may be multiple objects in an image.

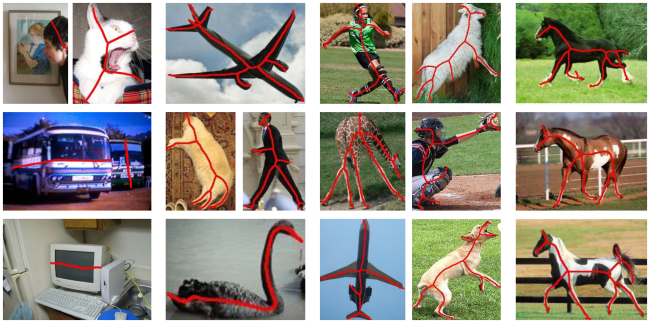
\begin{figure}[ht]
  \begin{overpic}[width=1\linewidth]{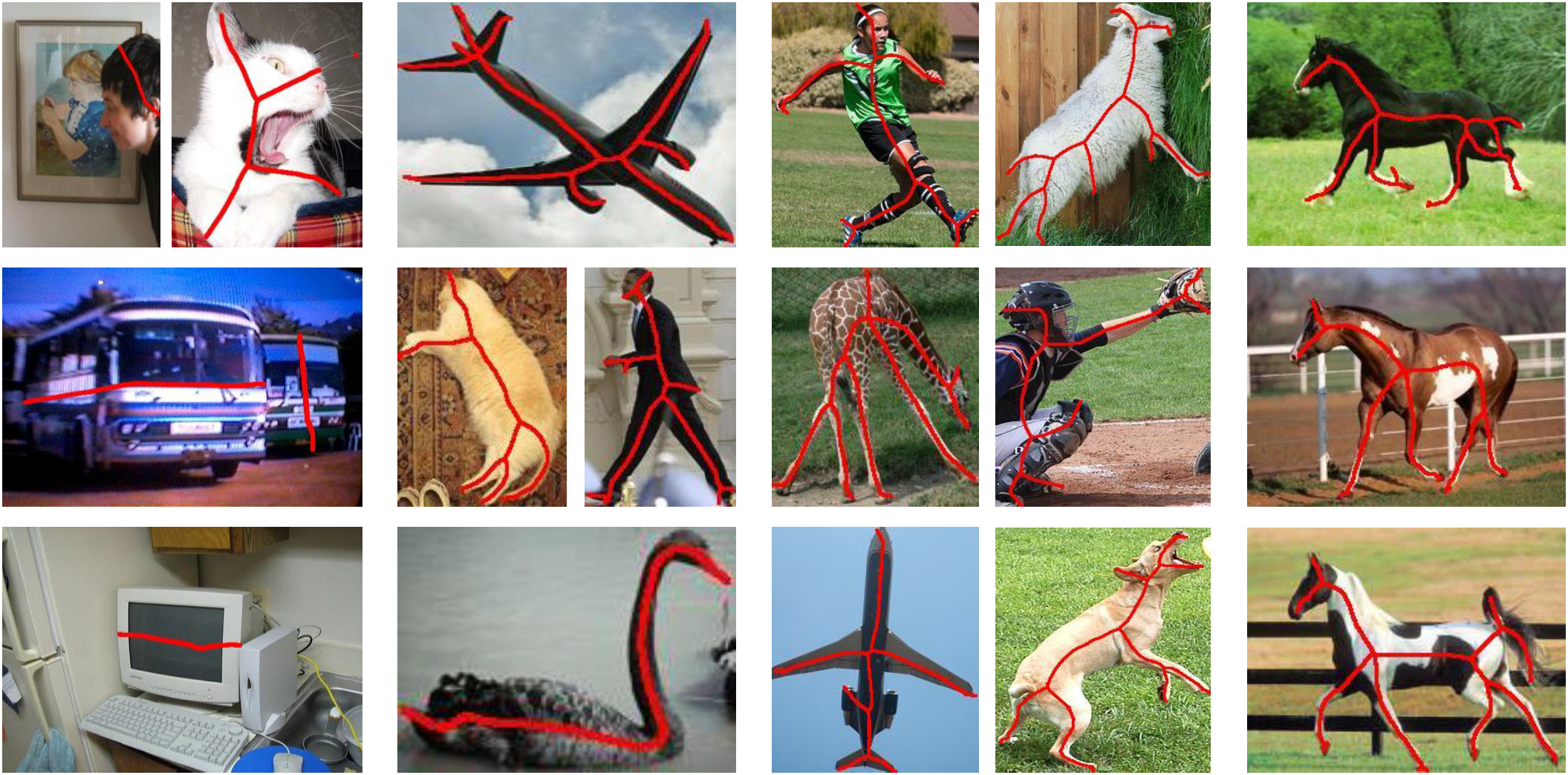}
  \put (1,-4) {\scriptsize{(a) SYM-PASCAL}}
  \put (26,-4) {\scriptsize{(b) SK-SMALL}}
  \put (54,-4) {\scriptsize{(c) SK-LARGE}}
  \put (79,-4) {\scriptsize{(d) WH-SYMMAX}}
  \end{overpic}\vspace{6pt}
  \caption{Example images and corresponding skeleton ground-truths 
  	(red curves) of several skeleton datesets.
  }\label{fig:dataset-examples}
\end{figure}

Some representative example images and corresponding skeleton ground-truths of these datasets are shown in \figref{fig:dataset-examples}.

\subsection{Implementation Details}
We implement the proposed architecture based on the openly available caffe~\cite{jia2014caffe} framework.
The hyper-parameters and corresponding values are:
base learning rate ($10^{-6}$),
mini-batch size (1),
momentum (0.9)
and maximal iteration (40000).
We decrease the learning rate every 10,000 iterations with factor 0.1.

We perform the same data augmentation operations with FSDS for fair comparison.
The augmentation operations are:
(1) random resize images (and gt maps) to 3 scales (0.8, 1, 1.2),
(2) random left-right flip images (and gt maps);
and (3) random rotate images (and gt maps) to 4 angles (0, 90, 180, 270).

\subsection{Evaluation Protocol}
The skeleton response map $\hat{Y}$ is obtained through \eqref{eq:sk-resp},
to which a standard non-maximal suppression (NMS) is then applied to obtain the thinned skeleton map for evaluation.
%
%
We evaluate the performance of the thinned skeleton map $\hat{Y}$ in terms of F-measure=
$2\frac{Precision \cdot Recall}{Precision + Recall}$
as well as the precision recall curve (PR-curve) w.r.t ground-truth $G$.
By applying different thresholds to $\hat{Y}$,
a series of precision/recall pairs are obtained to draw the PR-curve.
The F-measure is obtained under the optimal threshold over the whole dataset.

\subsection{Skeleton Detection}
\begin{figure*}[t]
  \centering  
  \includegraphics[width=1\textwidth]{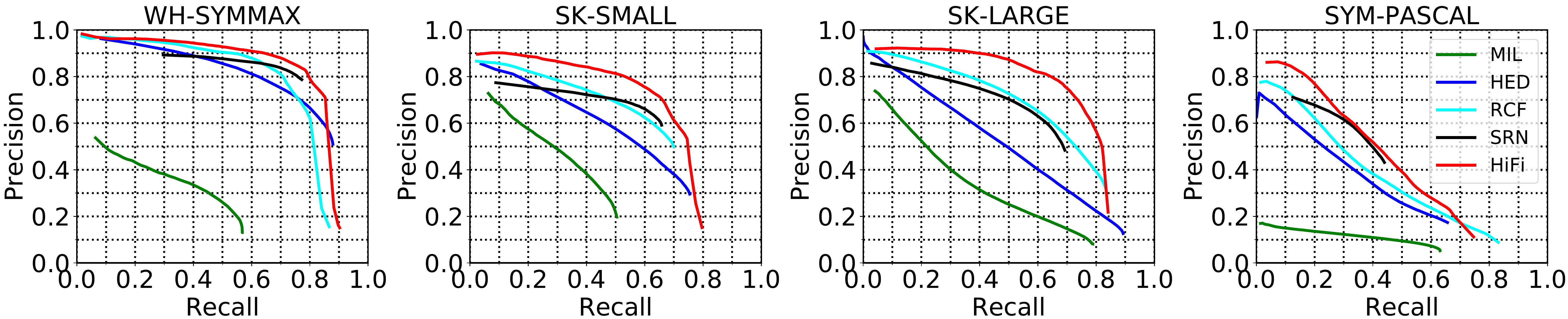}\\
  \vspace{-7pt}
  \caption{Precision Recall curves of recent CNN-based methods
    HED \protect\cite{xie2015holistically},
    RCF \protect\cite{liu2017richer},
    FSDS \protect\cite{shen2016object},
    SRN \protect\cite{ke2017side} and one learning based
    method MIL \protect\cite{tsogkas2012learning}.}\vspace{-6pt}
  \label{fig:pr-sk}
\end{figure*}

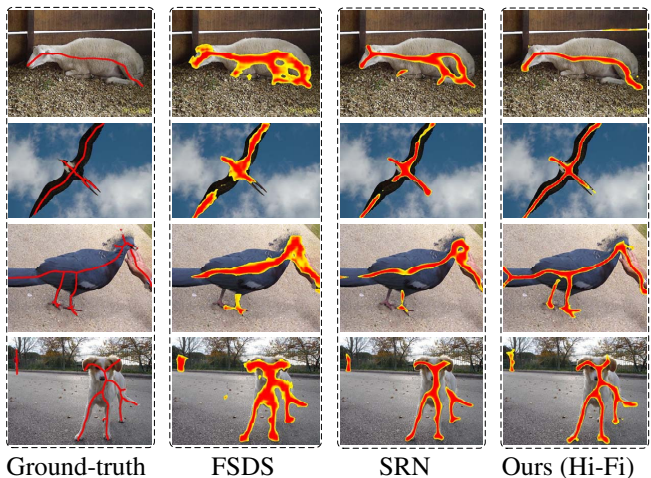
\begin{figure}[!h]
  \centering
  \begin{overpic}[width=1\linewidth]{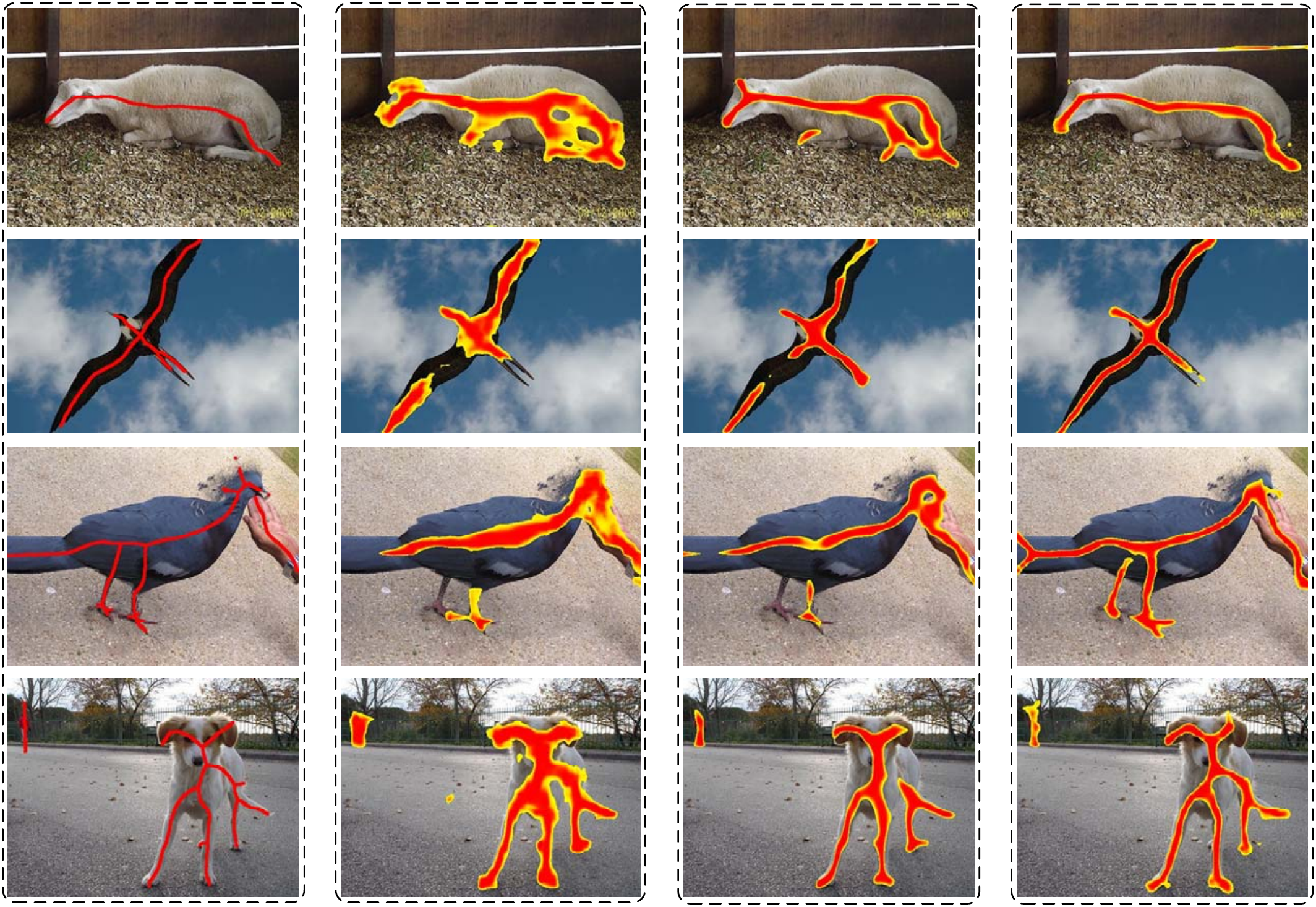}
   \put (0.2,-4){Ground-truth}
   \put (32,-4){FSDS}
   \put (58,-4){SRN}
   \put (77,-4){Ours (Hi-Fi)}
  \end{overpic}\vspace{4pt}
  \caption{
   Representative detection results.
   Our detected skeletons are more continuous and finer (thinner) than others.
   }\label{fig:sk-result-examples}\vspace{-10pt}
\end{figure}

We test our method on four aforementioned datasets.
Example images and ground-truths of these datasets are shown in \figref{fig:dataset-examples},
some detection results by different approaches are shown in ~\figref{fig:sk-result-examples}.
Similar to RCF~\cite{liu2017richer} we perform multi-scale detection by
resizing input images to different scales (0.5, 1, 1.5) and average their results.
We compare the proposed approach with other competitors including
one learning based method MIL~\cite{tsogkas2012learning},
and several recent CNN-based methods: FSDS~\cite{shen2016object}, SRN~\cite{ke2017side},
HED~\cite{xie2015holistically}, RCF~\cite{liu2017richer}).
FSDS and SRN are specialized skeleton detectors, 
HED and RCF are developed for edge detection.
Quantitative results are shown in \figref{fig:pr-sk} and Tab.\ref{tab:sk-fmeasure}, 
our proposed method outperforms the competitors in both terms of F-measure and PR-curve.
Some representative detection results are shown in \figref{fig:sk-result-examples}.

Comparison results in \tabref{tab:sk-fmeasure} reveal that 
with 1st-level hierarchical integration (Hi-Fi-1),
our method (Hi-Fi-1) already outperforms others with a significant margin.
Moreover, by integrating 1st-level integrated features we obtain 
the 2nd-level integration (Hi-Fi-2) and the performance
witnesses a further improvement (Limited by the GPU memory, we didn't implement Hi-Fi-3).
Architecture details of Hi-Fi-1 and Hi-Fi-2 are illustrated in 
\figref{fig:arch-fsds-srn-hifi2-h}.
%

%

\newcommand{\tabincell}[1]{\begin{tabular}{@{}c@{}}#1\end{tabular}}   

\begin{table}[!h]
  \centering
  \setlength\tabcolsep{6.4pt}
  \begin{tabular}{l|c|c|c|c}
    \hline
    Methods & \tabincell{WH-\\SYMMAX} & \tabincell{SK-\\SMALL} &
        \tabincell{SK-L\\ARGE} & \tabincell{SYM-\\PASCAL} \\
    \hline
    MIL & 0.365  & 0.392 & 0.293 & 0.174 \\
    HED & 0.732  & 0.542 & 0.497 & 0.369 \\
    RCF & 0.751  & 0.613 & 0.626 & 0.392     \\
    FSDS& 0.769  & 0.623 & 0.633 & 0.418 \\
    SRN & 0.780  & 0.632 & 0.640 & 0.443 \\
    \hline
    \textbf{Hi-Fi} & \textbf{0.805} & \textbf{0.681} & \textbf{0.724} & \textbf{0.454} \\
    \hline
    \end{tabular}\vspace{-6pt}
  \caption{F-measure comparison between different methods
    on four popular skeleton datasets.
    Our proposed Hi-Fi network outperforms other methods
    with an evident margin.
  }\label{tab:sk-fmeasure}%
\end{table}%
\vspace{-10pt}
\subsection{Ablation Study} \label{sec:exp-analysis}
We do ablation study on SK-LARGE dataset to further probe the proposed method.
\vspace{-10pt}\paragraph{Different Feature Integrating Mechanisms.}
We compare different feature integrating mechanisms including:
(1) FSDS~\cite{shen2016object},
(2) SRN ~\cite{ke2017side} with \emph{deep-to-shallow} refinement,
(3) Hi-Fi-1 with 1 level hierarchical integration (~\figref{fig:arch-fsds-srn-hifi2-h} (Hi-Fi-1)),
and (4) Hi-Fi-2 with 2 level hierarchical integration (~\figref{fig:arch-fsds-srn-hifi2-h} (Hi-Fi-2)).
Results are shown in \tabref{tab:feat_inte}.\vspace{-2pt}
\begin{table}[!h]
  \centering
  \setlength\tabcolsep{15pt}
  \begin{tabular}{c|c|c|c} \hline
    FSDS & SRN  & \textbf{Hi-Fi-1} & \textbf{Hi-Fi-2} \\
    \hline
    0.633 & 0.640  & \textbf{0.703} & \textbf{0.724} \\
    \hline
    \end{tabular}\vspace{-6pt}
  \caption{Performance of different feature integration mechanisms.
  }\label{tab:feat_inte}\vspace{-2pt}
\end{table}%
\vspace{-20pt}
\paragraph{Hierarchical Integration versus Direct Fusing.}
To further justify the proposed \emph{hierarchical feature integration} mechanism (Hi-Fi),
we compare the proposed Hi-Fi network with another architecture
described in \figref{fig:arch-feat-fuse} (c), which fuses features from all levels together at once.

\begin{figure}[!h]
\centering
  \includegraphics[width=0.8\linewidth]{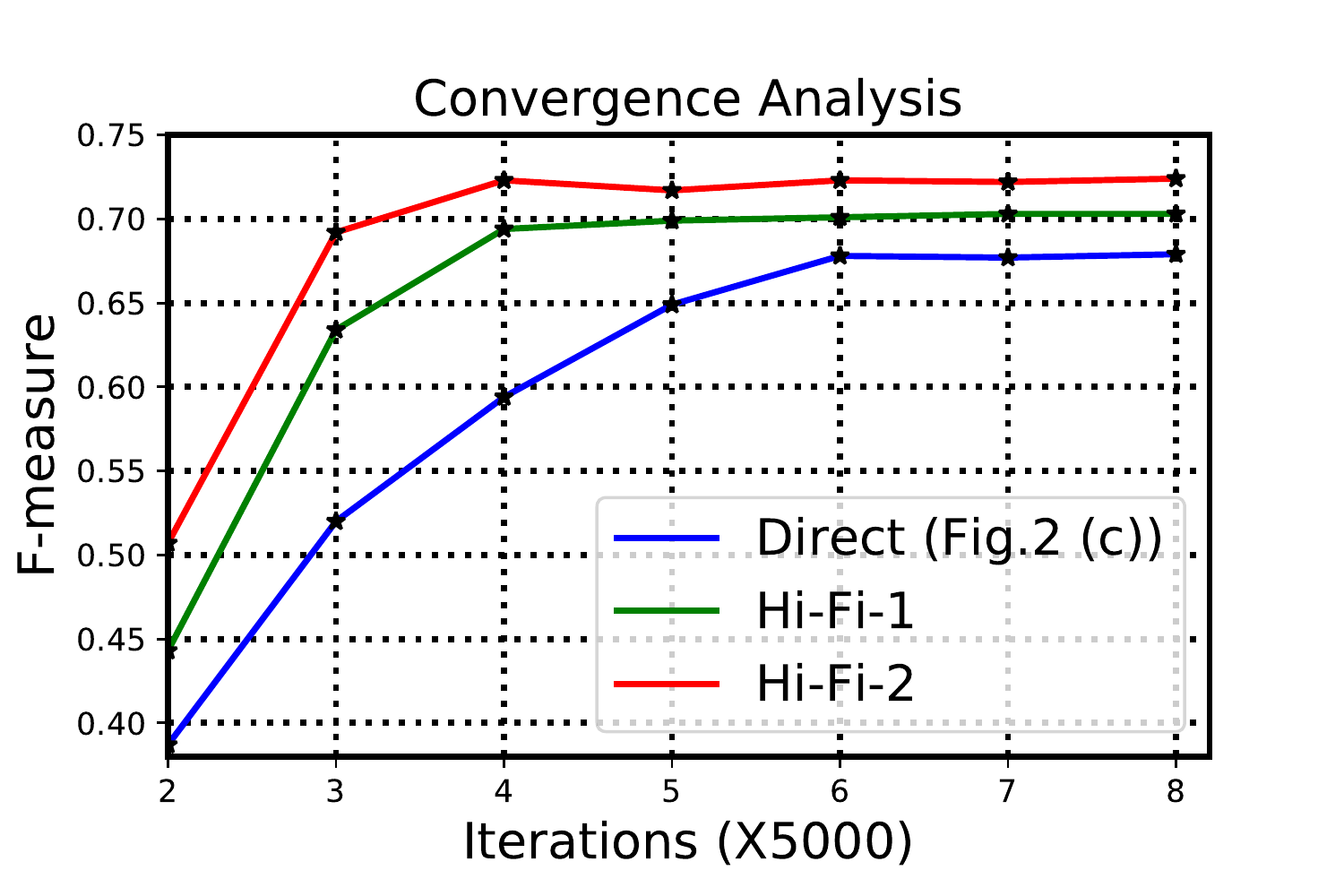}\vspace{-8pt}
  \caption{
    Hierarchical Feature Integration (Hi-Fi) versus Direct Fusing (\figref{fig:arch-feat-fuse} (c)).
  }\vspace{-8pt}
  \label{fig:convergence}
\end{figure}

The comparison results shown in \figref{fig:convergence} support our claim that ``learning an integration of nearby feature levels is easier than learning combination of features of all levels", as evidenced by a faster convergence and better converged performance.
\vspace{-10pt}

\vspace{-2pt}\paragraph{Integrating $K$ Feature Levels at Each Step.}
We also test models that fuse $K$ ($K=1,2,...,5$) consecutive feature levels at each step, 
and the results are summarized in \tabref{tab:k-levels}.
When $K=1$ the model reduces to FSDS where side-outputs are
working independently.
$K=2$ represents our proposed Hi-Fi (Hi-Fi-1 and Hi-Fi-2) which
combines every two nearby feature levels,
and $K=5$ is identical to \figref{fig:arch-feat-fuse} (c) that combines all feature levels at once.
\begin{table}[htbp]
  \centering
  \renewcommand{\arraystretch}{1}
  \setlength\tabcolsep{6.5pt}
  \begin{tabular}{c|c|c|c|c|c} \hline
    $K$ & 1      &  2     & 3      & 4     & 5 \\
    \hline
    F-measure & 0.633 & 0.703  & 0.689  & 0.690 & 0.679   \\
    \hline
    \end{tabular}\vspace{-4pt}
  \caption{Comparison of models that fuse $K$ feature levels at each step.
  }\label{tab:k-levels}\vspace{-10pt}
\end{table}%

\begin{figure}[htbp]
\centering
  \begin{overpic}[width=1\linewidth]{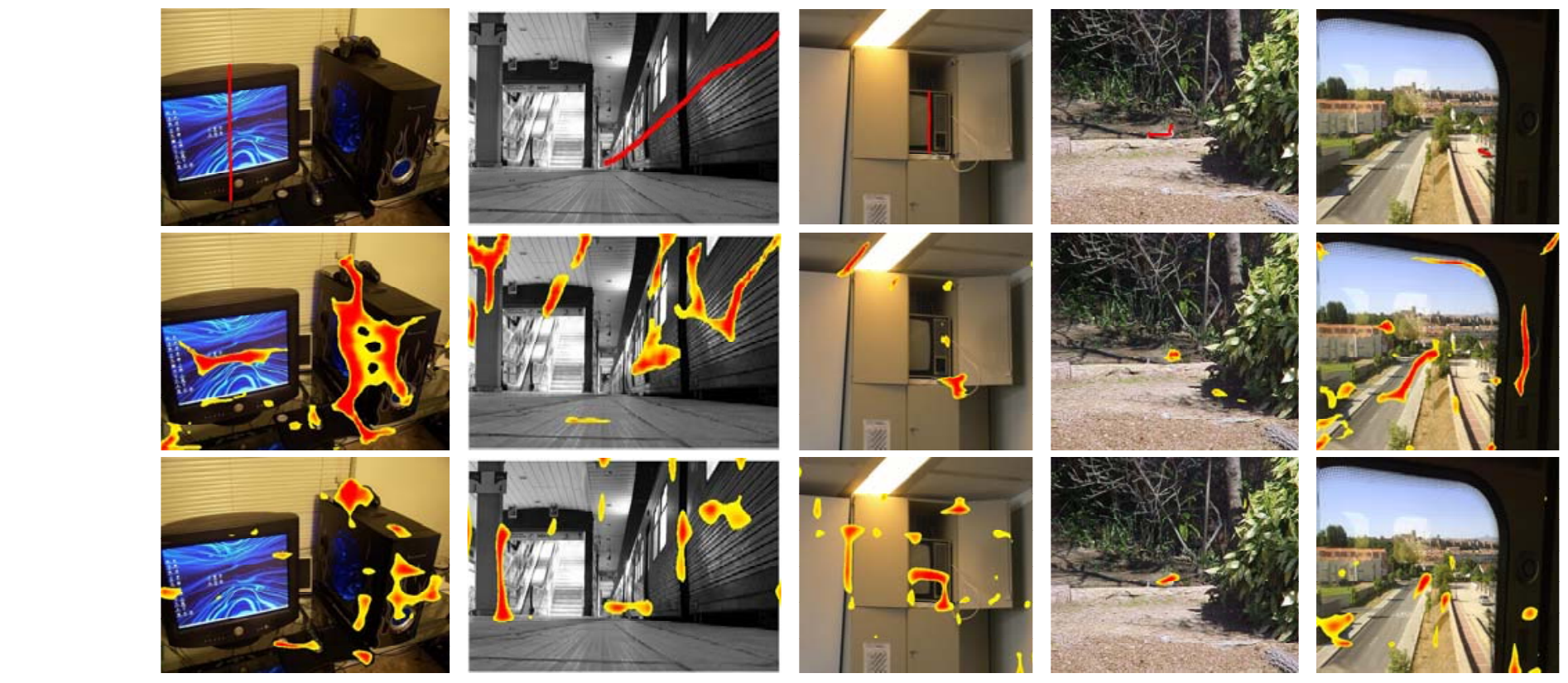}
  \put (1.2, 35){GT}
  \put (0, 20){Hi-Fi}
  \put (0.25, 6){SRN}
  \end{overpic}\vspace{-6pt}
  \caption{The top-5 worst detections on SYM-PASCAL dataset.
  	The results are ranked according to F-measure w.r.t ground-truths.
  }\vspace{-2pt}
  \label{fig:failure-case}
\end{figure}

\vspace{-12pt}\paragraph{Failure Case Exploration.}
Since our method achieves the least performance gain on SYM-PASCAL~\cite{ke2017side}, we analysis
the failure cases on this dataset.
We select top-5 worst detections (ranked by F-measure w.r.t  ground-truths) shown in \figref{fig:failure-case}.
Failure cases on this dataset are mainly caused by the ambiguous annotations and
not-well selected objects.
Our method (and also others) cannot deal with `square-shaped' objects like monitors and doors whose skeletons are hard to define and recognise.

\section{Conclusion}
We propose a new CNN architecture named Hi-Fi for skeleton detection.
Our proposed method has two main advantages over existing systems:
(a) it enables mutual refinement with both \emph{deep-to-shallow} and
\emph{shallow-to-deep} guidance;
(b) it recursively integrates nearby feature levels and supervises all intermediate
integrations, which leads to a faster convergence and better performance.
Experimental results on several benchmarks demonstrate that our method significantly
outperforms the state-of-the-arts with a clear margin.

\section*{Acknowledgments}
This research was supported by NSFC (NO. 61620106008, 61572264, 61672336),
Huawei Innovation Research Program,
and Fundamental Research Funds for the Central Universities.

\bibliographystyle{named}
\bibliography{HiFiSkeleton}
\end{document}